%% file: rojas.tex
\documentclass[letterpaper, 10 pt, conference]{ieeeconf}  
\IEEEoverridecommandlockouts                               
\usepackage{graphicx}                       
\usepackage{graphics}                       
\usepackage{epsfig}                         
\usepackage[tight,footnotesize]{subfigure}  

\usepackage{amssymb,amsmath}
\usepackage{mdwmath}
\usepackage{mdwtab}
\usepackage{commath}                        
\usepackage{eqparbox}
\input{rap2texdefs}    

\usepackage{stfloats}                       

\usepackage[T1]{fontenc}

\title{\LARGE \bf Robot Contact Task State Estimation via Position-based Action Grammars.}
\author{Juan Rojas, Zhengjie Huang, and Kensuke Harada. \\
\thanks{Juan Rojas is with the School of Electromechanical Engineering in the Guangdong University of Technology in Guangzhou, China.}%
\thanks{Zhengjie Huang is with the School of Software at Sun Yat Sen University in Guangzhou, China.}%
\thanks{Kensuke Harada is with the Graduate School of Engineering Science at Osaka University in Osaka, Japan.}%
}
\begin{document}
\maketitle
\thispagestyle{empty}
\pagestyle{empty}
\bstctlcite{IEEEexample:BSTcontrol}
\begin{abstract}
Uncertainty is a major difficulty in endowing robots with autonomy. Robots often fail due to unexpected events.
In robot contact tasks are often design to empirically look for force thresholds to define state transitions in a Markov chain or finite state machines. Such design is prone to failure in unstructured environments, when due to external disturbances or erroneous models, such thresholds are met, and lead to state transitions that are false-positives.
The focus of this paper is to perform high-level state estimation of robot behaviors and task output for robot contact tasks.
Our approach encodes raw low-level 3D cartesian trajectories and converts them into a high level (HL) action grammars. Cartesian trajectories can be segmented and encoded in a way that their dynamic properties, or ``texture'' are preserved. Once an action grammar is generated, a classifier is trained to detect current behaviors and ultimately the task output.
The system executed HL state estimation for task output verification with an accuracy of $86\%$, and behavior monitoring with an average accuracy of: $72\%$. The significance of the work is the transformation of difficult-to-use raw low-level data to HL data that enables robust behavior and task monitoring. Monitoring is useful for failure correction or other deliberation in high-level planning, programming by demonstration, and human-robot interaction to name a few. The data, code, and supporting resources for this work can be found at: https://www.juanrojas.net/research/position\_based\_action\_grammar.
\end{abstract}
\section{INTRODUCTION}\label{sec:Intro}
%
%
Uncertainty is a major difficulty in endowing robots with autonomy. Robots often fail due to unexpected events or behaviors. Low-level state estimation using robot pose has been extensively studied to help robots improve their belief about their state \cite{2008Springer-Siciliano-HandbookRobotics-EstimationCh4}. However, for contact tasks the interpretation of signals is not as straightforward. One key question is how can we use information from an FT sensor to know the system state? Especially at a high level (i.e. a behavior description) where the knowledge can be exploited to correct failures or deliberate. In practice, contact tasks are often designed to empirically look for force thresholds to define state transitions in a Markov chain or finite state machine. Such design is prone to failure in unstructured environments, when due to external disturbances or erroneous models, the force thresholds are met but lead to state transitions that are false-positives. Consider an assembly task that consists of an alignment and insertion. If the two parts do not mate properly during the alignment, the force threshold might nonetheless still be met, giving rise to the insertion behavior, but in which case the insertion will fail. The focus of this paper is to perform continuous state estimation (also referred to as process monitoring). In particular, we are interested in yielding high level state estimates that are useful representations for error correction and action deliberation.

As stated earlier, process monitoring has undertaken discrete event detection approaches. In such cases, contact points are detected and then evaluated as a contact sequence \cite{1998IJRR-Hovland-HMM_ProcessMonitorAsmbly, 2010CASE-Rodriguez-FailureDetAsmbly_ForceSigAnalysis, 2011IROS-Rodriguez-AbortRetry, 2015ICRA-Golz-TactileSensingLearnContactKnowledge}. With respect to continuous detection, our previous work \cite{2013IJMA-Rojas-TwrdsSnapSensing,2014ICRA-Rojas-EarlyFC, 2014ICMA-Luo-SnapSVM}, used a hierarchical taxonomy that encoded relative change from FT signatures and increasingly abstracted basic primitives to perform state estimation of a snap assembly task and identified behavior and anomaly information using statistical methods. In \cite{2013IROS-DiLello-BayesianContFaultDetection}, a sticky-Hierarchical Dirichlet Process Hidden Markov Model was used for continuous estimation of alignment tasks for a block and identified class anomalies caused by a various extraneous objects.

Additionally, we have begun to consider the inclusion of multi-sensor modalities for contact task's continuous state estimation. The neuroscience literature has been showing evidence on how humans perform haptic tasks best when using multi-sensory inputs \cite{2015FrHumNS-Whitwell-RTVision_Tact_Visual_Agnosia_RmHapticFB, 2007HFE-Morris-HaptFbEnhancesForceSkillLearning}.
While in this paper, we do not yet attempt a multi-model system, we wish to test the efficacy of performing high level continuous state estimation of contact tasks with \emph{position information} while using the same principle used earlier in the wrench domain--that is, encoding of relative changes to yield high-level representations. Our hypothesis is that information from the position domain will corroborate and complement the results derived from state estimation in the wrench domain.
\begin{figure}[htb]
    \centering
        \includegraphics[width=8cm,height=6cm]{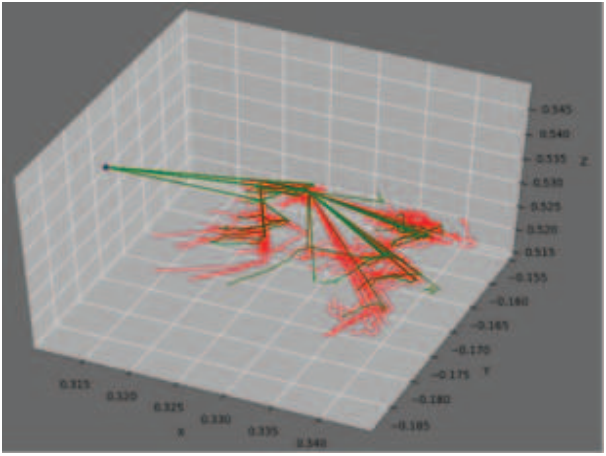}
        \caption{A series of 3D plots representing snap assembly tasks under controlled failure experiments.}
        \label{fig:failureTraj}
\end{figure}

In the area of process monitoring using position data, we find two approaches: Iterative 3D chain code construction, and Neighborhood feature extraction with a multi-level representation. The former uses two adjacent vectors with which, an alphabet of motion directions is generated. For neighborhood feature extraction there are two main lines of work: (i) the use of Frenet frames (FF) instead of 3D chain code constructs \cite{2012MICCAI-Ahmidi-PathPlannSurgeonSkillEval,2013MICCAI-Ahmidi-StringMotifDescrToolMotion_SkillGestures}, and (ii) the Scaled Indexing of General Shapes (S-IGS) \cite{2014CASE-Yang-ScaleIndexGralShapesMtnRec,2009PR-Wu-FlexibleSignDescr_AdapMtnTrajReps,2011IROS-Yang-InvarTrajIndexRT3DMtnRec}, which analyzes curvature and torsion at each trajectory point to do point-level or segment-level descriptions and construct primitives from point sequences with similar features. S-IGS methods have only been applied to human hand motions, while FF methods have been applied to suturing task for minimally invasive surgery.

In our work, we wish to analyze whether FF-based approaches can effectively perform high level state estimation for contact tasks. In particular, we tested snap assemblies, which are characterized by high frequency contact during alignment and insertion phases, see Fig. \ref{fig:failureTraj} as an example. Such application would be novel, since the aforementioned approaches have not been tested in highly non-smooth, noisy signals. We want to study how accurately these approaches can perform both \emph{behavior} (or sub-task) recognition and \emph{task} (output) verification. If a force-dominant task can be accurately estimated with position data, then we can complement state estimation methods that use wrench information as the sensor modality. The contribution of our work is an efficacy analysis in use of FF related techniques for state monitoring and task verification for robot contact tasks. Our analysis looks at both sub-and-whole-task identification, in success and failures cases, both in simulation and with a real humanoid robot.

Our approach can be organized into three stages: segmentation, encoding, and classification. Signal segmentation is effected through the use of FF-based methods. The encoding of the segmentation uses Direct Curve Coding (DCC), which gives rise to a grammar of motions, and, the action grammar classification is done using linear SVM.  The system executed HL state estimation for task output with an average value of $86\%$, and behavior monitoring with an average performance value of: $72\%$. The significance of the work is the transformation of difficult-to-use raw low-level data to HL data that enables robust behavior and task monitoring. Monitoring is useful for failure correction or other deliberation in high-level planning, programming by demonstration, and human-robot interaction to name a few.
\section{Methodology}\label{sec:methodology}
\begin{figure*}[bt!]
    \centering
        \includegraphics[scale=2.5]{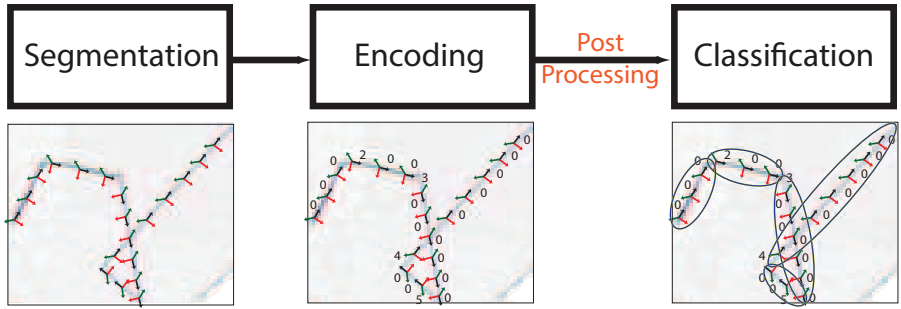}
        \caption{Methodology overview: 3D curves can be segmented, encoded, and trained for behavior and task classification.}
        \label{fig:overview}
\end{figure*}

High-level state estimation using position data is bootstrapped by segmenting a 3D position curve according to FF or AFF (see Sec. \ref{subsec:FF}), the segmented curve is then encoded using DCC. Encoded signals of different lengths, are post-processed to have equal number of segments. Post-processed data is then trained using classification algorithms (both for behavior estimation and task output verification). An overview of the process is presented in Fig. \ref{fig:overview}.
\subsection{Segmentation and Encoding}\label{subsec:FF}
The use of FFs is motivated by the fact that this discretization is able to capture the ``texture'' of the motion in a way other techniques have not \cite{2012MICCAI-Ahmidi-PathPlannSurgeonSkillEval}. Texture is captured since FFs capture the motion's local curvature from the point of view of an observer traveling on the curve. This also makes the technique coordinate-independent and robot setup-independent. This principle is similar to the approach taken in our state estimation work using wrench information where we capture relative change by fitting the signal with straight line segments using regression over a growing window \cite{2013IJMA-Rojas-TwrdsSnapSensing}. Both focus on relative change. FFs consists of a base frame consisting of three orthonormal vectors: tangent $\vec{t}$, normal $\vec{n}$, and binormal $\vec{b}$. On continuous curves $s(t)=\int_{0}^{t}\norm{ r'(\sigma)} d\sigma$, the vectors are defined according to Eqtn. 1.
\begin{eqnarray}
  \nonumber     \vec{t} &=& \frac{dr}{ds},             \\
                \vec{n} &=& \frac{dt/ds}{ \norm{dT/ds} },   \\
  \nonumber     \vec{b} &=& \vec{t}\times \vec{n}.
  \label{eqtn:contFF}
\end{eqnarray}
For a discrete curve connected by a sequence of points $x_k(k=1,2,3,...)$, discrete Frenet frames (DFF) assign the orthonormal vectors according to Eqtn. 2.
\begin{eqnarray}
  \nonumber     \vec{t_k} &=& \frac{\vec{x}_{k+1} - \vec{x}_k}{\norm{ \vec{x}_{k+1}-\vec{x}_k} },                 \\
                \vec{n_k} &=& \frac{ \vec{t}_{k-1}\times \vec{t}_k}{\norm{ \vec{t}_{k-1} \times \vec{t}_k} },    \\
  \nonumber     \vec{b_k} &=& \vec{b}_k \times \vec{t}_k.
  \label{eqtn:discFF}
\end{eqnarray}
The first frame placement on the curve is arbitrary and selected by the user. Each succeeding frame is placed every unit step and follows Eqtn's 1 or 2. Direction can be encoded iteratively as long as a current $x_s$ and a previous $x_{s-1}$ frames exist. The sequence of FFs, generated from the position time-series, are encoded into a string representation, (hereto referred as an action grammar) by mapping direction changes through a small set of canonical directions.

The number of canonical directions is related to the number of vectors used in the frame assignment. The base case, with three orthonormal vectors, can represent a minimal set of directions. Namely: forward, backward, up, down, left, right, and no motion. These directions can be encoded with strings: ``0,1,...,6'' respectively. Mathematically, the direction $d$ for time-step $s$ is selected as the tendency of motion towards one of the 7 possible directions. $d$ is assigned by projecting the current tangent vector $t^s$, onto the previous orthonormal basis $x_{s-1}$ and choosing the direction that $t^s$ is \emph{closest} to one of the existing directions.
\begin{gather}\label{eqtn:directions}
    \nonumber   d^s=\arg\max_i \{t^s.[x_{s-1}] \},                      \\
                x_s=[t^{s-1},n^{s-1},b^{s-1},-t^{s-1},-n^{s-1},-b^{s-1}].
\end{gather}
This encoding technique is known as Direct Curve Coding (DCC).
\subsubsection{Accumulated Frenet Frames (AFF)}\label{subsubsec:AFF}
\begin{figure}[t!]
    \centering
        \includegraphics[scale=0.75]{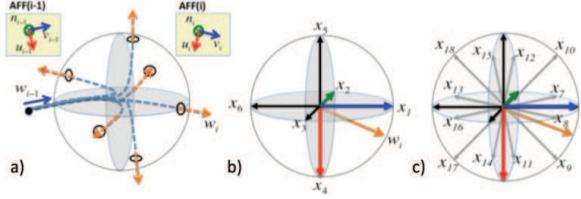}
        \caption{Illustration of possible direction changes in a curve and possible space partitioning configurations.  (a) Shows the change in direction between succeeding frames, (b) DCC7 with 7 vectors and a directional threshold of $\pi/4$, (c) DCC19 with 19 vectors and a directional threshold of $\pi/8$. \cite{2013MICCAI-Ahmidi-StringMotifDescrToolMotion_SkillGestures}.}
        \label{fig:spacePartitions}
\end{figure}

One weakness in the FF method is that it is unable to adequately represent smooth trajectories, where the curve is characterized by small direction changes. The FF's action grammar only updates when turns greater than $pi/4$ take place. AFF's, on the other hand, is a method that accumulates change over small spatial or temporal windows and is more sensitive to gradual change. It does so first by expanding DCC to be systematic and allow different granularity levels to partition space. DCC partitions space depending on a base parameter $p$ according to Eqtn. \ref{eqtn:dcc}:
\begin{gather}\label{eqtn:dcc}
    [\vec{x}]_p = \left[ \vec{x} \right]_{p-1} \bigcup \left[ \frac{\vec{x}_a + \vec{x}_b}{ \norm{ \vec{x_a + x_b}} } \right], \\ \nonumber
    \abs{ \vec{x}_a \otimes \vec{x}_b } \neq 0,                                                                       \\
    \mbox{g.t.} \quad \vec{x}_a,\vec{x}_b \in \left[ [\vec{x}]_{p-1} \right].                                                    \nonumber
\end{gather}
This base case, where $p=1$, yields 7 vectors: 6 orthogonal (up, down, forward, back, left, right) + one null vector for null motion. It divides the space into octants. Granularity can be increased (and thus the set size of canonical directions) by inserting more vectors in the projection function (Eqtn. 3). Additional vectors can be inserted by taking the union of the previous base $x_{p-1}$ with the normalized sum of its parts. The partitioning level ascribed to a base $p=2$ yields 19 vectors. The third base has 91 vectors, and the fourth 2891. Note that the space is not evenly partitioned in these two latter cases. Fig. \ref{fig:spacePartitions} illustrates such space partitions.

\begin{figure}[b!]
    \centering
        \includegraphics[scale=0.3]{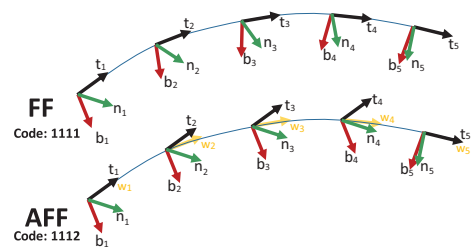}
        \caption{Comparison between action grammar updates between FF and AFFs.}
        \label{fig:aff}
\end{figure}

Secondly, in AFF, new frames no longer set their $\vec{t}_s$ vector along the current tangent of the curve (represented as $\vec{w}_s$ in Fig. \ref{fig:aff}), instead, new frames keep the same orientation until a directional threshold is met. At the time, we align the new $\vec{t}_s$ with the current tangent $\vec{w}_s$. In Fig. \ref{fig:aff}, we compare FF and AFF. Notice that both techniques keep the same code for the first four steps, but for the fifth step, the directional change of the curve has accumulated enough that the directional threshold is met and AFF updates its direction. In doing so, the action grammar is also modified.
Directional thresholds are set by the equation: $\alpha_p=\pi/2^{p+1}$. The equation is dependent on the DCC base number. For DCC7, the threshold is $\alpha_1=\pi/4$. For DCC19 it is $\alpha=\pi/8$. Once the directional threshold is met, we use the same projection function from Eqtn. \ref{eqtn:directions}, although with the appropriate number of vectors for a DCC specification. Possible direction changes and DCC specifications can be appreciated in Fig. \ref{fig:spacePartitions}.
In the end, the action grammar--a sequence of labels that span the entire trajectory, describes the local curvature changes experienced by the 3D position curve. This grammar encodes behavior and task properties that help us characterize how the motion was executed. In effect we have taken a continuous position signal and mapped into discrete changes that will aid in the high-level estimation of the robot contact task.
\subsection{Classification}\label{subsec:classification}
In this work, we classify behaviors and task output. A task is composed of a sequence of behaviors. A behavior's action grammar is obtained by performing DCC on the time segment of a given behavior. At this time, the time transitions are provided by the robot controller: when a threshold is met, a transition time is recorded. A task's action grammar is the set of string labels for the entire task. The action grammar is used as a feature vector for the supervised classification algorithm. Fig. \ref{fig:labelMap} shows an encoded color map for the action grammar of a full snap assembly task for multiple trials. The map provides an intuitive way to see the texture and patterns across a task.

As part of the classification, feature vectors used across trials need to have the same length for proper evaluation. In this work we take two approaches to align the features: one is a blind approach where there is information loss and the second approach is a simple interpolation procedure. In the blind approach, we search for the feature vector with the smallest length in an experiment and then proceed to cut the additional elements contained in the longer vectors for other trials through a logical AND operation.
In the resampling case, we consider all trajectories in a training or testing set; count the number of frame assignments for each trajectory, and compute the average number of points. We then roughly approximate the length of each curve by summing the linear distance between points and divide by the average number of points. This gives a unit length for resampling. All curves are then re-segmented with this length and later encoded with sDCC.
\subsection{Support Vector Machines}\label{subsubsec:svm}
Linear Support Vector Machines (SVMs) approximate a boundary to separate binary classes through a hyperplane for large feature spaces. The feature vector is used to learn a hyperplane: $\omega^T x − b=0$, where $\omega$ are the weights and $b$ is the bias from the zero point. In effect, the separation of each training point from the hyperplane is the functional margin $\hat{\gamma}^(i)$ and can be modeled as:
\begin{equation}\label{eqtn:hyperplane}
    \hat{\gamma}^{(i)} = y^{(i)}(\omega^{(i)}x + b)
\end{equation}
Here the pair ${y(i), x(i)}$ represents the class as $y^{(i)} \in {1,−1}$ and $x^{(i)}$ is the input vector
for training and testing. The SVM optimizes the functional margin by maximizing the distance to both true and
failure cases by solving the quadratic programming problem:
\begin{gather}\label{eqtn:gamma}
  \max (\gamma)                                       \nonumber       \\
  \quad \mbox{ s.t.} \quad \gamma = \min_{i=1,..,m} \hat{\gamma},
\end{gather}
where, $\gamma$ is the geometrical margin of the input points from the hyperplane. The larger the geometrical margin the more
accurate the classifier. Our linear classifier was tested with a linear, polynomial, and a radial basis function as kernels using Scikit's machine learning library \cite{scikit-learn}.
\section{Experiments}\label{sec:Experiments}
\begin{figure}[b!]
    \centering
        \includegraphics[scale=0.5]{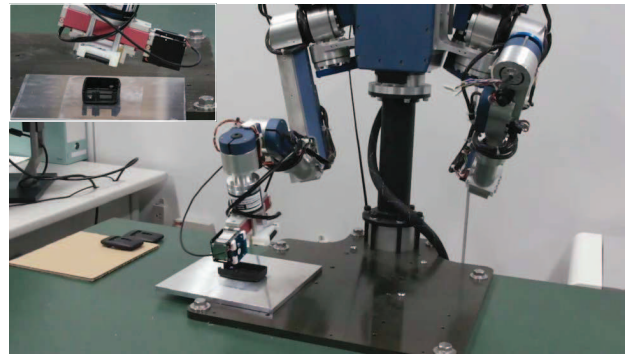}
        \caption{The HIRO-NX robot performing the snap assembly of a male and female camera part with 4 cantilever snaps. Top-left is a zoomed-in version of the assembly.}
        \label{fig:expSetUp}
\end{figure}

NX-HIRO, a 6 DoF dual-arm anthropomorph robot was run both in the OpenHRP 3.0 simulation environment \cite{2004IJRR:Kaneheiro:OpenHRP} and in the real robot. Male and female 4-snap cantilever camera parts were used. The male part was rigidly mounted on the robot's wrist, while the female snap was rigidly fixed to the ground as in Fig. \ref{fig:expSetUp}. For this work we consider the Pivot Approach strategy presented in \cite{2013IJMA-Rojas-TwrdsSnapSensing} where, a successful assembly task consists of four behaviors: an approach, an alignment, an insertion, and a mating. The Approach behavior drives the male part along a smooth trajectory until it contacts the female part at an angle at a docking pivot as in Fig. \ref{fig:expSetUp}. The rest of the behaviors are achieved using modular force-moment controllers as stated in \cite{2013IJMA-Rojas-TwrdsSnapSensing}.

With respect to the test configuration, two kinds of experiments were conducted. One experiment only considered controlled failure experiments, while the other only considered successful assemblies. The failure experiments introduced small but large enough deviations from a nominal trajectory in the Approach state to generate failure early in the assembly task. The small deviations are characteristic of what a human adult would make when trying to enact a parts alignment but narrowly misses the mark. The failure experiment was only conducted in simulation, classification was conducted for both behaviors and task output. The other successful assembly task experiments, simulation and real-robot tests were conducted, and behaviors and task output were both classified. The data sets stand as follows:
\begin{itemize}
  \item A: Successful Assembly Simulation
  \item B: Controlled Failed Assembly Simulation
  \item C: Successful Assembly with the Real-Robot
\end{itemize}

As for our testing, we sought to thoroughly analyze the performance of the system. We performed classification evaluations for data sets A,B,C, as well as their combinations: AB, AC, and ABC. We ran linear SVM with three kernel types: linear (SVC based on libsvm \& LinearSVC based on liblinear), polynomial, and RBF. A 2-to-20 cross fold validation was used--we start with 2 folds and move up to 20. For each $k$-value we run classification 10 times. Average, maximum, and minimum classification accuracy results are collected. We also compared the results from FF and AFFs both running DCC19.
\subsection{Results} \label{sec:results}
\begin{figure*}[tb!]
    \centering
        \includegraphics[scale=1.1]{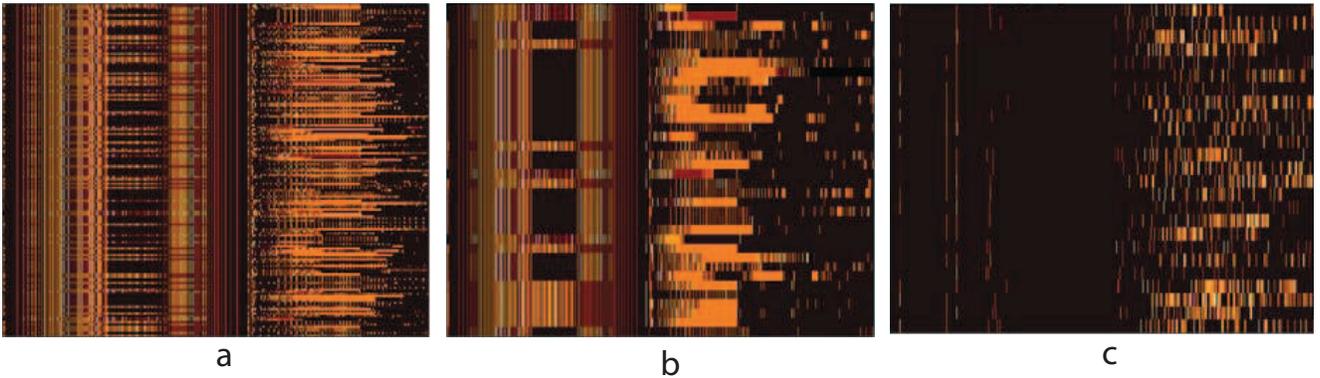}
        \caption{Encoded color map for action grammars. The x-axis represents color coded string labels from the action grammar for a contact snap assembly task under controlled failures. The y-axis represents different trials. The visible patterns in the map indicate similarities across different parts of the trajectory. (a) and (b) show different controlled failure experiments in simulation, (c) illustrates the successful assembly task by the real robot.}
        \label{fig:labelMap}
\end{figure*}

We first present color-coded maps for the action grammars of some of the data sets in Fig. \ref{fig:labelMap}. The color-coded maps, provide an intuitive representation of motion texture as well as the patterns therein. Each row corresponds to an encoded string for a given trajectory in the data subset and the 19-element alphabet (given by the DCC19-encoded strings) is mapped to 19 levels of color brightness.
From the color coded maps in Fig. \ref{fig:labelMap}, notice the similarities within class for successful and failure trials. There is more variability in the failure class because in these controlled experiments there are up to 6 different subclasses (that we did not classify in this work). Not explicitly considering the failure sub-classes hurts our classification result and this is something that will be addressed in future work.

The first set of results to report is the SVM performance for different kernels. SVM performance is reported both for behavior classification and task output classification. The numbers reported are the average of all data set results (A,B,C,AB,AC,ABC) for both FF and AFF:
\begin{table}
    \centering
        \caption{SVM Performance with Different Kernels.}\label{tbl:SVMPerformance}
        {\setlength{\tabcolsep}{2pt} \setlength{\extrarowheight}{2pt}%
            \begin{tabular}{c|c|c|c|c|c}
                        &               &       SVC\_Linear  & Linear\_SVC  & RBF       & Polynomial        \\
                \hline
                FF      &   behavior    &       72          & 60            & 48        & 66                \\
                        &   task        &       86          & 85            & 63        & 81                \\
                \hline
                AFF     &   behavior    &       70          & 59            & 44        & 66                \\
                        &   task        &       86          & 85            & 63        & 80                \\
            \end{tabular}
        }
\end{table}
Table \ref{tbl:SVMPerformance} reveals that SVC\_Linear performs best across the board: both under FF and AFF for both behavior and task classification.

The next set of results presents the overall performance of SVC\_Linear under 2-to-20 fold cross validation, for all data sets under different alignment methods for both behavior and task output, for both FF and AFF. The results can be seen in Table \ref{tbl:overalFF}.
\begin{table}
    \centering
        \caption{Overall Performance for the FF method.}\label{tbl:overalFF}
        {\setlength{\tabcolsep}{3pt} \setlength{\extrarowheight}{2pt}%
            \begin{tabular}{c|c|c|c|c|c|c|c|c|c}
            Method  & Level & Alignment &   & A & B & C & AB & AC & ABC \\
             \hline
                    &           &           & avg & 81 & 76 & 64 & 64 & 62 & 64 \\
                    &           & Interp    & min & 73 & 73 & 51 & 54 & 57 & 58 \\
                    & Behavior  &           & max & 93 & 80 & 73 & 76 & 69 & 70 \\
             \cline{3-10}
                    &           &           & avg & 83 & 77 & 82 & 76 & 68 & 67 \\
                    &           & Cut       & min & 69 & 76 & 77 & 71 & 65 & 63 \\
              FF    &           &           & max & 95 & 82 & 90 & 82 & 72 & 72 \\
             \cline{2-10}
                    &           &           & avg &   & 78 & 92  & 89  & 80 & 83 \\
                    &           & Interp    & min &   & 71 & 86  & 84  & 70 & 77 \\
                    & Task      &           & max &   & 83 & 100 & 96  & 85 & 90 \\
             \cline{3-10}
                    &           &           & avg &   & 77 & 100 & 100 & 81 & 80 \\
                    &           & Cut       & min &   & 74 & 100 & 100 & 78 & 75 \\
                    &           &           & max &   & 85 & 100 & 100 & 85 & 86 \\
            \hline

                    &           &           & avg &    & 76 & 64 & 64 & 63 & 64 \\
                    &           & Interp    & min &    & 73 & 56 & 51 & 56 & 56 \\
                    & Behavior  &           & max &    & 80 & 72 & 76 & 71 & 70 \\
             \cline{3-10}
                    &           &           & avg &    & 77 & 82 & 76 & 70 & 68 \\
                    &           & Cut       & min &    & 74 & 77 & 72 & 65 & 65 \\
              AFF   &           &           & max &    & 81 & 87 & 81 & 74 & 71 \\
             \cline{2-10}
                    &           &           & avg &   & 78 & 91  & 88  & 81 & 82 \\
                    &           & Interp    & min &   & 72 & 85  & 83  & 76 & 77 \\
                    & Task      &           & max &   & 88 & 97  & 96  & 85 & 87 \\
             \cline{3-10}
                    &           &           & avg &   & 76 & 99  & 100 & 80 & 81 \\
                    &           & Cut       & min &   & 71 & 99  & 100 & 75 & 77 \\
                    &           &           & max &   & 82 & 100 & 100 & 85 & 88 \\
            \end{tabular}
        }
\end{table}

With respect to ``Alignment'' methods, Table \ref{tbl:overalFF} shows that generally, the cut-off alignment method performs better than the resampling alignment. The results can be misleading as there is information loss in the blind approach. Even so, the resampling approach performed well for task classification when using AFF-DCC19. For the successful assembly with the real-robot, it correctly classified $91\%$ of the time, while for data set combinations AB, AC, and ABC, it correctly classified between $81\%-88\%$ of the time.

With respect to the performance between output task classification and behavior classification, the former has a much higher accuracy on most data set combinations except for B, the failure data set. We mentioned in Sec. \ref{sec:Experiments} that the failure data can be subdivided in multiple dissimilar classes, hence the poorer classification in sets with failure data: B, BC and ABC. The task-level classification has an outstanding 100\% classification accuracy on the success data set combinations C and AB with cut-off alignment. We see that the longer the action grammar, the higher classification accuracy. Behavior classification rated as follows: $64\%-81\%$ for FF-Interpolated for classes A, B, C. FF-Cut does better from $77\%-83\%$. AFF-Interpolated for classes A, B, C, performed less accurately: $64\%-76\%$, while FF-Cut did better: $76\%-82\%$.

With respect to the performance between the FF and AFF techniques (both using DCC19), we generally did not see improvements anywhere. In fact, FF seems to do marginally better all around. The reason for this might be that in fact our contact task problem consists of trajectories full of discontinuities. Therefore, the advantages offered by AFF for smooth curves cannot be appreciated in this domain.
\section{DISCUSSION} \label{sec:Discussion}
In this work, we performed high level state estimation of robot contact tasks by taking 3D Cartesian trajectories and encoding them in action grammar and then classifying them to monitor behavior and task output. Our trajectories, unlike previous works, have segments with significant non-smoothness and discontinuities. Nonetheless, the results are comparable to previous works that analyzed suturing tasks in minimally invasive surgery. In our domain, the system had average classification accuracy of $86\%$ for output task verification. Though the system also reached values of $92\%-100\%$ for real-robot success assemblies. For behavior HL state estimation the average classification rate was: $72\%$ but reached $82\%$ for real-robot successful assemblies (the average value was lowered by not considering the classification of failure subclasses). In \cite{2013MICCAI-Ahmidi-StringMotifDescrToolMotion_SkillGestures}, for their DCC19 experiments on data set DS-I, with known state boundaries and for spatial quantization, their average behavior classification rate was: $82.16\%$. For their second data set DS-II, for spatial DCC19, their average classification results across k-fold validation and one-user out validation (for all their skill sets) was $75.22\%$.

Our classification results may be also hurt due to the fact that the feature vector fed to the SVM classifier lacks notions of similarity between different trials. A similarity measure should be designed and used to formulate a new feature vector for the classifier. The second limitation is that our classifier is only running offline. We are considering implementing a dynamic time warping solution that is a powerful similarity measure that is optimal for online monitoring \cite{2013VLDB-Toyoda-PatternDiscoveryDataStreams_TimeWarping} and combine this with online supervised methods or probabilistic models.

Additionally, we are not yet performing automatic detection of state transitions. We currently rely on transition times provided by the controller, which we earlier stated to be an important source of error.

Finally, we are interested in integrating this work into a multi-modal architecture for HL state estimation. By combining both low-level and high-level information from multi-sensor modalities promises to increase robustness in state estimation, especially for contact tasks, that consist of harder to predict signals.
\section{CONCLUSION} \label{sec:Conclusion}
This work presented a system to perform HL state estimation for behavior monitoring and task verification for robot contact tasks. The system generates an action grammar from a 3D position trajectory generated from the robot end-effector that encodes the dynamic properties of the motion. This grammar is used as a feature vector for a classifier that in turns helps to identify executing behaviors or the final result of a task. The work is significant in that it transforms raw low-level data into useful high-level that can be used for failure correction or deliberation about future actions.
\section{Acknowledgements} \label{sec:Acknowledgements}
This work is supported by ``Major Project of the Guangdong Province Department for Science and Technology (2014B090919002), (2016B0911006).''
\bibliographystyle{IEEEtran}
\bibliography{\jobname}
\end{document}

%% file: rap2texdefs.tex
\usepackage{amsmath,amsfonts,amssymb}

\newcommand{\beq}{\begin{equation}}
\newcommand{\eeq}{\end{equation}}
\newcommand{\bear}{\begin{eqnarray}}
\newcommand{\bears}{\begin{eqnarray*}}
\newcommand{\eear}{\end{eqnarray}}
\newcommand{\eears}{\end{eqnarray*}}
\newcommand{\bdm}{\begin{displaymath}}
\newcommand{\edm}{\end{displaymath}}
\newcommand{\lba}{\left[\begin{array}}
\newcommand{\ear}{\end{array}\right]}

%% file: rojas.bbl
\begin{thebibliography}{10}
\providecommand{\url}[1]{#1}
\csname url@samestyle\endcsname
\providecommand{\newblock}{\relax}
\providecommand{\bibinfo}[2]{#2}
\providecommand{\BIBentrySTDinterwordspacing}{\spaceskip=0pt\relax}
\providecommand{\BIBentryALTinterwordstretchfactor}{4}
\providecommand{\BIBentryALTinterwordspacing}{\spaceskip=\fontdimen2\font plus
\BIBentryALTinterwordstretchfactor\fontdimen3\font minus
  \fontdimen4\font\relax}
\providecommand{\BIBforeignlanguage}[2]{{%
\expandafter\ifx\csname l@#1\endcsname\relax
\typeout{** WARNING: IEEEtran.bst: No hyphenation pattern has been}%
\typeout{** loaded for the language `#1'. Using the pattern for}%
\typeout{** the default language instead.}%
\else
\language=\csname l@#1\endcsname
\fi
#2}}
\providecommand{\BIBdecl}{\relax}
\BIBdecl

\bibitem{2008Springer-Siciliano-HandbookRobotics-EstimationCh4}
B.~Siciliano and O.~Khatib, \emph{Springer handbook of robotics, Ch. 4 Sensing
  and Estimation}.\hskip 1em plus 0.5em minus 0.4em\relax Springer Science \&
  Business Media, 2008.

\bibitem{1998IJRR-Hovland-HMM_ProcessMonitorAsmbly}
G.~E. Hovland and B.~J. McCarragher, ``Hidden markov models as a process
  monitor in robotic assembly,'' \emph{The International Journal of Robotics
  Research}, vol.~17, no.~2, pp. 153--168, 1998.

\bibitem{2010CASE-Rodriguez-FailureDetAsmbly_ForceSigAnalysis}
A.~Rodriguez, D.~Bourne, M.~Mason, G.~F. Rossano, and J.~Wang, ``Failure
  detection in assembly: Force signature analysis,'' in \emph{Automation
  Science and Engineering (CASE), 2010 IEEE Conference on}.\hskip 1em plus
  0.5em minus 0.4em\relax IEEE, 2010, pp. 210--215.

\bibitem{2011IROS-Rodriguez-AbortRetry}
A.~Rodriguez, M.~T. Mason, S.~S. Srinivasa, M.~Bernstein, and A.~Zirbel,
  ``Abort and retry in grasping,'' in \emph{Intelligent Robots and Systems
  (IROS), 2011 IEEE/RSJ International Conference on}.\hskip 1em plus 0.5em
  minus 0.4em\relax IEEE, 2011, pp. 1804--1810.

\bibitem{2015ICRA-Golz-TactileSensingLearnContactKnowledge}
S.~Golz, C.~Osendorfer, and S.~Haddadin, ``Using tactile sensation for learning
  contact knowledge: Discriminate collision from physical interaction,'' in
  \emph{Robotics and Automation (ICRA), 2015 IEEE International Conference
  on}.\hskip 1em plus 0.5em minus 0.4em\relax IEEE, 2015, pp. 3788--3794.

\bibitem{2013IJMA-Rojas-TwrdsSnapSensing}
J.~Rojas, K.~Harada, H.~Onda, N.~Yamanobe, E.~Yoshida, K.~Nagata, and Y.~Kawai,
  ``Towards snap sensing,'' \emph{International Journal of Mechatronics and
  Automation}, vol.~3, no.~2, pp. 69--93, 2013.

\bibitem{2014ICRA-Rojas-EarlyFC}
J.~Rojas, K.~Harada, H.~Onda, N.~Yamanobe, E.~Yoshida, and K.~Nagata, ``Early
  failure characterization of cantilever snap assemblies using the pa-rcbht,''
  in \emph{IEEE International Conference on Robotics and Automation (ICRA)},
  2014, pp. 3370--3377.

\bibitem{2014ICMA-Luo-SnapSVM}
W.~Luo, J.~Rojas, T.~Guan, K.~Harada, and K.~Nagata, ``Cantilever snap
  assemblies failure detection using svms and the rcbht,'' in
  \emph{Mechatronics and Automation (ICMA), 2014 IEEE International Conference
  on}, Aug 2014, pp. 384--389.

\bibitem{2013IROS-DiLello-BayesianContFaultDetection}
E.~Di~Lello, M.~Klotzbucher, T.~De~Laet, and H.~Bruyninckx, ``Bayesian
  time-series models for continuous fault detection and recognition in
  industrial robotic tasks,'' in \emph{Intelligent Robots and Systems (IROS),
  2013 IEEE/RSJ International Conference on}.\hskip 1em plus 0.5em minus
  0.4em\relax IEEE, 2013, pp. 5827--5833.

\bibitem{2015FrHumNS-Whitwell-RTVision_Tact_Visual_Agnosia_RmHapticFB}
R.~L. Whitwell, T.~Ganel, C.~M. Byrne, and M.~A. Goodale, ``Real-time vision,
  tactile cues, and visual form agnosia: removing haptic feedback from a
  "natural" grasping task induces pantomime-like grasps,'' \emph{Frontiers in
  human neuroscience}, vol.~9, 2015.

\bibitem{2007HFE-Morris-HaptFbEnhancesForceSkillLearning}
\BIBentryALTinterwordspacing
D.~Morris, H.~Tan, F.~Barbagli, T.~Chang, and K.~Salisbury, ``Haptic feedback
  enhances force skill learning,'' in \emph{Proceedings of the Second Joint
  EuroHaptics Conference and Symposium on Haptic Interfaces for Virtual
  Environment and Teleoperator Systems}, ser. WHC '07.\hskip 1em plus 0.5em
  minus 0.4em\relax Washington, DC, USA: IEEE Computer Society, 2007, pp.
  21--26. [Online]. Available: \url{http://dx.doi.org/10.1109/WHC.2007.65}
\BIBentrySTDinterwordspacing

\bibitem{2012MICCAI-Ahmidi-PathPlannSurgeonSkillEval}
N.~Ahmidi, G.~D. Hager, L.~Ishii, G.~L. Gallia, and M.~Ishii, ``Robotic path
  planning for surgeon skill evaluation in minimally-invasive sinus surgery,''
  in \emph{International Conference on Medical Image Computing and
  Computer-Assisted Intervention}.\hskip 1em plus 0.5em minus 0.4em\relax
  Springer, 2012, pp. 471--478.

\bibitem{2013MICCAI-Ahmidi-StringMotifDescrToolMotion_SkillGestures}
N.~Ahmidi, Y.~Gao, B.~B{\'e}jar, S.~S. Vedula, S.~Khudanpur, R.~Vidal, and
  G.~D. Hager, ``String motif-based description of tool motion for detecting
  skill and gestures in robotic surgery,'' in \emph{Medical Image Computing and
  Computer-Assisted Intervention--MICCAI 2013}.\hskip 1em plus 0.5em minus
  0.4em\relax Springer, 2013, pp. 26--33.

\bibitem{2014CASE-Yang-ScaleIndexGralShapesMtnRec}
J.~Yang, H.~Xu, X.~Zhou, and Y.~F. Li, ``Scaled indexing of general shapes for
  complicated 3d motion recognition,'' in \emph{2014 IEEE International
  Conference on Automation Science and Engineering (CASE)}, Aug 2014, pp.
  236--241.

\bibitem{2009PR-Wu-FlexibleSignDescr_AdapMtnTrajReps}
S.~Wu and Y.~F. Li, ``Flexible signature descriptions for adaptive motion
  trajectory representation, perception and recognition,'' \emph{Pattern
  Recognition}, vol.~42, no.~1, pp. 194--214, 2009.

\bibitem{2011IROS-Yang-InvarTrajIndexRT3DMtnRec}
J.~Yang, Y.~Li, and K.~Wang, ``Invariant trajectory indexing for real time 3d
  motion recognition,'' in \emph{2011 IEEE/RSJ International Conference on
  Intelligent Robots and Systems}.\hskip 1em plus 0.5em minus 0.4em\relax IEEE,
  2011, pp. 3440--3445.

\bibitem{scikit-learn}
F.~Pedregosa, G.~Varoquaux, A.~Gramfort, V.~Michel, B.~Thirion, O.~Grisel,
  M.~Blondel, P.~Prettenhofer, R.~Weiss, V.~Dubourg, J.~Vanderplas, A.~Passos,
  D.~Cournapeau, M.~Brucher, M.~Perrot, and E.~Duchesnay, ``Scikit-learn:
  Machine learning in {P}ython,'' \emph{Journal of Machine Learning Research},
  vol.~12, pp. 2825--2830, 2011.

\bibitem{2004IJRR:Kaneheiro:OpenHRP}
F.~Kanehiro, H.~Hirukawana, and S.~Kajita, ``Openhrp: Open architecture
  humanoid robotics platform,'' \emph{Intl. J. of Robotics Res.}, vol.~23,
  no.~2, pp. 155--165, 2004.

\bibitem{2013VLDB-Toyoda-PatternDiscoveryDataStreams_TimeWarping}
\BIBentryALTinterwordspacing
M.~Toyoda, Y.~Sakurai, and Y.~Ishikawa, ``Pattern discovery in data streams
  under the time warping distance,'' \emph{The VLDB Journal}, vol.~22, no.~3,
  pp. 295--318, 2013. [Online]. Available:
  \url{http://dx.doi.org/10.1007/s00778-012-0289-3}
\BIBentrySTDinterwordspacing

\end{thebibliography}
